\documentclass[letterpaper, 10 pt, conference]{ieeeconf}  

\IEEEoverridecommandlockouts

\usepackage{flushend}
\usepackage{balance} 

\let\labelindent\relax

\usepackage{etoolbox} 
\usepackage[utf8]{inputenc} 
\usepackage[T1]{fontenc}    
\usepackage[english]{babel} 
\usepackage[protrusion=true,expansion=true]{microtype}
\usepackage{comment}
\usepackage{cancel}
\usepackage{csquotes}
\usepackage{listings}
\usepackage{nameref}
\usepackage{paralist}
\usepackage{xspace}
\usepackage{setspace}
\usepackage{makeidx}
\usepackage{pdfpages}

\usepackage{amsmath,amssymb} 
\usepackage{amsfonts}       
\usepackage{mathtools}
\usepackage{array} 
\usepackage{nicefrac}       
\usepackage{breqn}          
\usepackage{textcomp}
\usepackage{bm} 
\usepackage{pifont}
\usepackage[
  separate-uncertainty = true,
  multi-part-units = repeat
]{siunitx}

\usepackage{graphicx}
\usepackage{adjustbox} 
\usepackage{epsfig}

\usepackage{float}
\usepackage{subcaption}
\usepackage[font=small,labelfont=bf]{caption}
\usepackage{lscape}      

\usepackage{tikz}  
\usepackage{makecell}
\usepackage{overpic}
\usepackage{algorithm}
\usepackage[noend]{algpseudocode}

\usepackage{array} 
\usepackage{tabularx}
\usepackage{multirow}
\usepackage{multicol}
\usepackage{booktabs}   

\usepackage{paralist}
\usepackage{enumitem}
\setlist[itemize]{noitemsep,leftmargin=*,topsep=0in}
\setlist[enumerate]{noitemsep,leftmargin=*,topsep=0in}

\usepackage{url}            
\PassOptionsToPackage{table}{xcolor}
\usepackage{color,colortbl} 
\usepackage{soul}

\PassOptionsToPackage{capitalize, noabbrev}{cleveref}
\usepackage[pagebackref=false,breaklinks=true,colorlinks,urlcolor=blue,citecolor=blue,linkcolor=blue,bookmarks=false]{hyperref}

\makeatletter
\let\NAT@parse\undefined
\makeatother
\usepackage[numbers,sort&compress]{natbib}

\usepackage{blindtext}
\usepackage{bbm}

\usepackage{lipsum}

\usepackage{titlesec}
\titlespacing{\section}{0pt}{0.4\baselineskip}{0.25\baselineskip}
\titlespacing{\subsection}{0pt}{0.25\baselineskip}{0.15\baselineskip}
\titlespacing{\subsubsection}{0pt}{0.05\baselineskip}{0.03\baselineskip}

\renewcommand{\paragraph}[1]{\vspace{1em}\noindent\textit{#1} --}

\newcommand{\dataName}{Syn-TODD\xspace}
\newcommand{\modelName}{MVTrans\xspace}
\newcommand{\cmark}{\ding{52}}%
\newcommand{\xmark}{\ding{56}}%

\title{\LARGE \bf
MVTrans: \underline{M}ulti-\underline{V}iew Perception of \underline{Trans}parent Objects 
}

\author{
    Yi Ru Wang$^{\dagger\sharp\ast}$, %
    Yuchi Zhao$^{\ddagger\ast}$, %
    Haoping Xu$^{\dagger\ast}$, %
    Sagi Eppel$^{\dagger}$, %
    \\
    Al\`{a}n Aspuru-Guzik$^{\dagger}$, %
    Florian Shkurti$^{\dagger}$, %
    Animesh Garg$^{\dagger	\mp}$ 
    \thanks{ \noindent 
    $^{\dagger}$University of Toronto \& Vector Institute, 
    } 
    \thanks{$^{\ddagger}$University of Waterloo,
    $^{	\mp}$Nvidia}
    \thanks{$^{\sharp}$University of Washington, {\tt\small yiruwang@cs.washington.edu}}
    \thanks{$^\ast$ Authors contributed equally}
}

\usepackage{color,soul}

\definecolor{chartblue}{RGB}{21, 53, 98}

\begin{document}

\maketitle
\thispagestyle{empty}
\pagestyle{empty}

\begin{abstract}

Transparent object perception is a crucial skill for applications such as robot manipulation in household and laboratory settings. Existing methods utilize RGB-D or stereo inputs to handle a subset of perception tasks including depth and pose estimation. However, transparent object perception remains to be an open problem. In this paper, we forgo the unreliable depth map from RGB-D sensors and extend the stereo based method. Our proposed method, \modelName, is an end-to-end multi-view architecture with multiple perception capabilities, including depth estimation, segmentation, and pose estimation. Additionally, we establish a novel procedural photo-realistic dataset generation pipeline and create a large-scale transparent object detection dataset, \dataName, which is suitable for training networks with all three modalities, RGB-D, stereo and multi-view RGB. \url{https://ac-rad.github.io/MVTrans/}

\end{abstract}

\section{INTRODUCTION}

Transparent objects are prevalent in our daily lives, and their use spans household, laboratory, and industrial settings. However, the unique specular properties of transparent objects cause perception challenges, particularly in areas of depth estimation, segmentation, and pose estimation. Specifically, transparent objects differ from common objects in their ability to inherit visual properties of the background, as well as distort light rays and hence the depth modality of commodity RGB-D sensors, which operate on the assumption that objects have opaque lambertian surfaces \cite{keypose}. 

Existing methods have addressed transparent object perception challenges in two ways. RGB-D based depth completion methods \cite{ClearGrasp, seeingGlass, zhu2021rgbd} recover estimated transparent object depth from raw sensor depth and RGB features, and use the predicted depth for pose estimation tasks. Another stream of works \cite{keypose, pmlr-v164-kollar22a} skip the unreliable sensor depth and directly work on transparent object pose estimation using stereoscopic imagery. Bypassing the depth completion problem brings advantages over RGB-D based methods. Namely, it unifies models from multi-step pipelines into a single end-to-end model. These stereo-based models have higher capacity than build-in depth sensor algorithms to handle non-lambertian surface objects. 

\begin{figure}[t]
	\centering
	\includegraphics[width=0.8\linewidth]{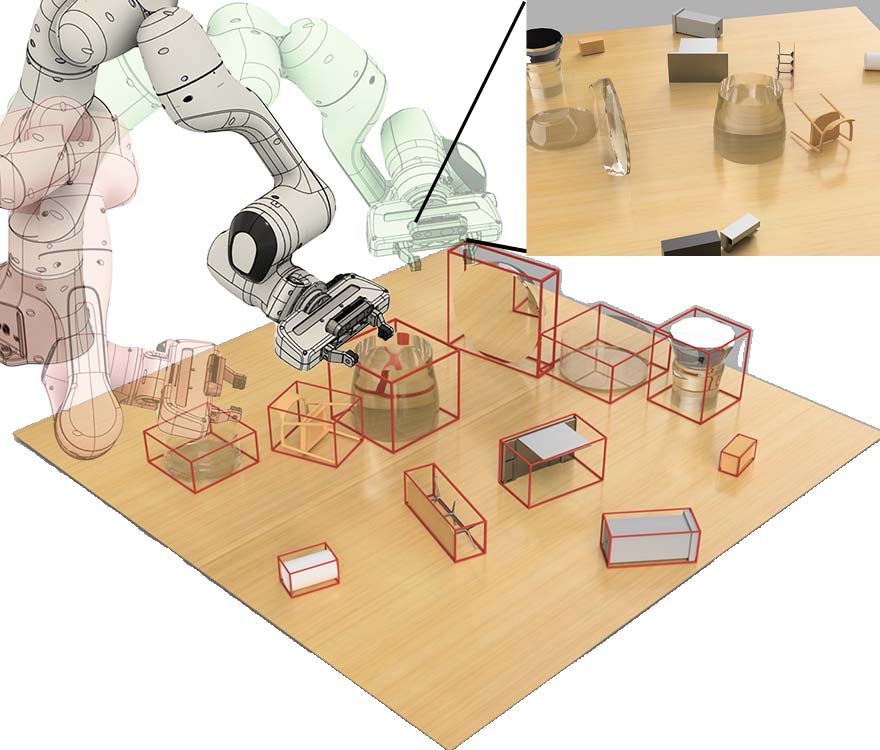} 
	\captionof{figure}{Given a set of multi-view images of the scene, the proposed method (\modelName) predicts segmentation mask, depth map, pose, and 3D bounding box for both opaque and transparent objects using an end-to-end multi-task perception network.}
	\label{fig:scene}
\end{figure}

\begin{figure*}[tp]
	\centering
    	\includegraphics[width=0.75\textwidth]{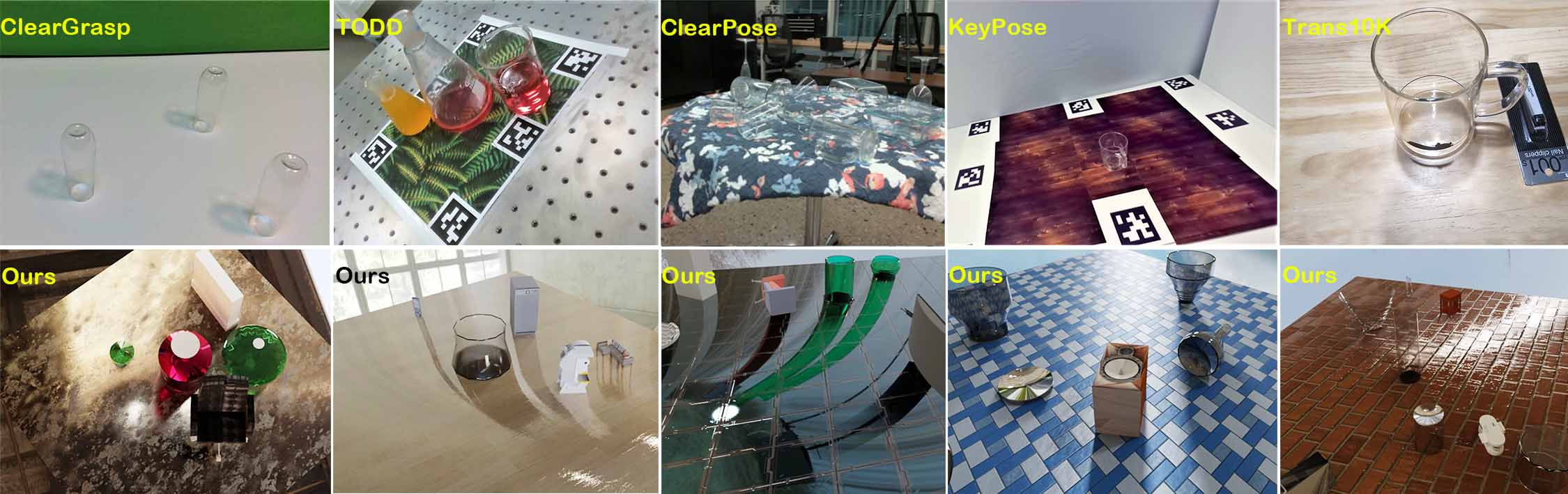} \\
	
	\captionof{figure}{\textbf{Sample images} of transparent object datasets. (a) Top row: existing methods including ClearGrasp \cite{ClearGrasp}, TODD \cite{seeingGlass}, ClearPose \cite{clearpose},   KeyPose \cite{keypose} and Trans10K \cite{trans10k}.  (b) Bottom row: Ours (\dataName). \dataName's scene complexity and object diversity are superior when compared with existing datasets.}
	\label{fig:dataset_compare}
\end{figure*}

Multi-view estimation extends stereo vision by providing richer information for a given scene especially in complex settings with occlusion. Multi-view methods \cite{ESPDepth, multiviewRGBD, Yao2018MVSNetDI} demonstrate superior performance in 3D vision tasks and have the potential to surpass stereo vision methods when handling transparency by enabling fusion of diverse viewing angles. This is especially convenient for autonomous settings, where an eye-in-hand camera can capture varying views to perceive and manipulate transparent objects \cite{seeingGlass,keypose}, shown in Figure \ref{fig:scene}. 

To train such multi-view networks, large-scale transparent object datasets are needed, yet existing ones are not suitable. Some works \cite{ClearGrasp, trans10k} focus on single view tasks and lack multi-view annotations. Other datasets collected using robots or SLAM  \cite{keypose, seeingGlass, clearpose} provide the multi-view images necessary for training. However, these real-world datasets are limited in object diversity and scene complexity.

In this work, we introduce an end-to-end multi-view architecture for perception of transparent objects, including depth estimation, segmentation, and scene understanding (pose and 3D bounding box prediction). Our method outperforms state-of-the-art RGB-D and stereo models on both real-world and synthetic datasets for all transparent object perception tasks. To ensure model generalizability, we present a novel pipeline for procedural photo-realistic dataset generation, and a large-scale transparent object dataset (\dataName). The dataset generation pipeline enables procedural generation of transparent objects, paired with domain randomized scene setup. In total, our dataset includes 1996 photo-realistic tabletop scenes with transparent and opaque objects, and 57 fully annotated views for each scene. In summary, our contributions are twofold:

\begin{enumerate}
  \item A novel end-to-end multi-view architecture, \modelName, for multi-task perception of transparent objects. It can perform  depth estimation, segmentation, and scene understanding including pose and 3D bound box prediction for every object in a given scene. Its performance exceeds current RGB-D and stereo-based methods.
  \item A large-scale transparent object dataset, \dataName. It has wide compatibility with RGB, RGB-D, stereo and multi-view based methods and superior scene complexity, object diversity and annotation richness. To create it, a rendering pipeline with domain randomization and procedural generation are employed.
\end{enumerate}

\section{Related Work}
\subsection{Transparent Object Perception}
Transparent objects pose additional problems for both 2D and 3D perception tasks, mainly due to its non-opaque, non-lambertian surface which breaks traditional methods' assumption of opaqueness \cite{keypose}. Similar to opaque object perception, transparent object perception tasks include segmentation, depth estimation, 3D bounding box and pose prediction.  

For \textbf{segmentation}, \cite{1467548, 1640865, https://doi.org/10.48550/arxiv.2003.13948, https://doi.org/10.48550/arxiv.2103.13279, DBLP:journals/corr/abs-2103-15734} aim to improve the performance of general object recognition and segmentation on particular transparent objects. \cite{DBLP:journals/corr/abs-2107-03172, https://doi.org/10.48550/arxiv.2003.13948, DBLP:journals/corr/abs-2103-15734,DBLP:journals/corr/abs-2101-08461} segment transparent and reflective surfaces for visual navigation and scene understanding. Several monocular RGB methods leverage the unique difference in appearance and texture along the edge of transparent vessels by incorporating boundary cues \cite{ https://doi.org/10.48550/arxiv.2003.13948, https://doi.org/10.48550/arxiv.2103.13279, DBLP:journals/corr/abs-2103-15734}. 

\textbf{Depth completion} is particularly important for RGB-D based networks when handling transparent objects, whose depth appears distorted when captured raw. Some works demonstrate the effectiveness of a global optimization approach, which leverages the combination of predicted surface normal, occlusion boundary and original depth for depth estimation guidance \cite{ClearGrasp, rgbd_depth}.  Other methods use encoder-decoder or Generative Adversarial Networks (GANs) to generate the completed depth map by regression \cite{senushkin2021decoder, RGBDGAN}.  

\textbf{3D BBox and pose prediction} methods concern either axis-aligned bounding box or oriented bounding box. Some existing works use CNNs to train from  RGB-D input \cite{Wang2019DenseFusion6O,He2020PVN3DAD,Wada2020MoreFusionMR}, where the depth is generated from aforementioned depth completion modules. Stereo vision methods avoid the distorted depth and directly take stereo image pairs as input. KeyPose \cite{keypose} is a stereo RGB and keypoint based method, and SimNet \cite{pmlr-v164-kollar22a} is a stereo and oriented 3D BBox based method. Our proposed \modelName architecture is capable of preforming all three perception tasks: segmentation, depth completion and scene understanding which includes pose and 3D BBox for every object in the scene.
\begin{figure*}[t!]
	\centering
	\includegraphics[width=0.85\textwidth]{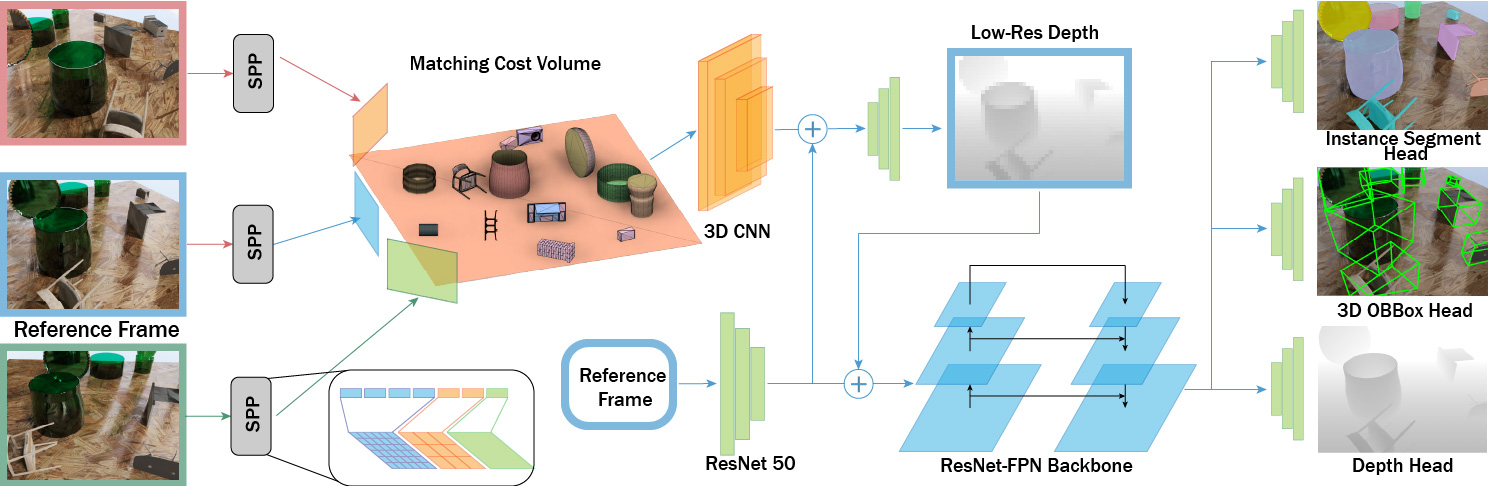} 
	\captionof{figure}{\textbf{\modelName Architecture.} Images from different viewing angles are used as input, in which one frame is selected as the reference frame for predictions. For each image, a shared spatial pyramid pooling (SPP) module extracts its features. Subsequent plane-sweeping warps are applied to non-reference frames to build the matching volume, which is then regularized by a 3D CNN. In parallel, the reference frame's 2D context features is extracted by ResNet-50. Concatenated 3D matching volume and 2D context features are used to generate a low-res depth prediction. Both RGB and depth features are fed to a ResNet-FPN backbone as well as downstream output heads, which predicts instance segmentation, 3D OBB and full resolution depth map.}
	\label{fig:architecture}
\end{figure*}
\subsection{Multi-view Perception} 
Single-view perception refers to estimation of scene parameters and properties using a single monocular image input. Multi-view perception refers to the broad category of using more than one view to infer 3D information from the captured scene. Some works impose the epipolar constraint, which leverages a pair of stereo images to learn the associated disparity and depth \cite{keypose, pmlr-v164-kollar22a}. Other works use supervisory signals for depth estimation guidance \cite{multiviewRGBD}, or enforce constraints regarding the spatio-temporal consistency between consecutive frames \cite{ESPDepth}. Recent works also demonstrate the advantage of plane sweeping volume algorithm for in multi-view 3D feature fusion \cite{Yao2018MVSNetDI, pmlr-v164-kollar22a}. Our work is within the broad multi-view perception category, in which we leverage multiple overlapping views for multi-task learning.

\subsection{Transparent Object Dataset}
Transparent objects lack an ideal benchmark dataset for 3D perception tasks. Existing transparent datasets come with different limitations. Trans10K \cite{trans10k} is a 2D segmentation dataset that consists of real images of transparent objects created by manual annotation. ClearGrasp \cite{ClearGrasp} proposed a synthetic 3D dataset with 9 unique objects. KeyPose \cite{keypose} collects a 3D real-world dataset of 15 unique objects using an eye-in-hand robot with only single object scenes. TODD \cite{seeingGlass} automates the collection and annotation process using eye-in-hand camera and AprilTags, its dataset includes 8 unique objects and complex scenes with cluttered and filled glassware. Recently, ClearPose \cite{clearpose} proposed to use SLAM and manual CAD model alignment for large-scale real world transparent object dataset creation. To improve these limitations, we create \dataName, a large-scale transparent dataset with superior scene complexity, object diversity, and annotation richness which supports model training with RGB-D, stereo and multi-view modalities, a sample of which is shown in Figure \ref{fig:dataset_compare}.

\section{\modelName: Multi-Task Model Architecture}

Our proposed method, \modelName, as shown in \autoref{fig:architecture}, is an end-to-end multi-view architecture with multiple perception capabilities, namely depth estimation, segmentation, and scene understanding (pose and 3D bounding box prediction for every opaque and transparent object in a given scene). \modelName takes a set of multi-view images as input, whose 2D features are back-projected into a 3D matching volume. The matching volume and reference image's 2D features are used to predict a low resolution depth map. Both the RGB features and low-res depth feature then passed through a backbone feature extractor before reaching the multi-task prediction heads for segmentation, 3D bounding box and pose estimation, and depth prediction.

\subsection{Local and Global Context Fusion for Multi-view Input}
Given a collection of RGB input \{$I_{1},..., I_{N}$\}, where $N \geq 2$ is the number of multi-view images, we refer to the first image as the reference image, and the others as support images whose captured view partially overlaps with the reference image scene. Each image has dimensions $H \times W \times 3$. Similar to the feature extraction process in \cite{long2021multiview}, expert networks are used to produce a Matching Volume (3D) and a Context Volume, which are jointly fused into a cost volume that encapsulates the local and global features. 

\textbf{Matching volume.} Each multi-view image is passed through a spatial pyramid pooling (SPP) module \cite{Chang2018PyramidSM} to encapsulate and aggregate context across different scales and locations to form feature maps. The feature maps of the support images are back-projected into the coordinate system of the reference image, at a stack of parallel planes with depths sampled based on ranges observed in the dataset. The goal of the back-projection is to capture and incorporate the photo-consistency of the warped images on the pixel level. At each sampled depth $z_d$, where $d \in [1,D]$ is the plane index, a planar homography transformation is applied to obtain the coordinate mapping from the support to the reference image: 

\begin{equation}
    \resizebox{0.5\linewidth}{!}{
    $
    u'_i \sim \mathbf{K} [R_r | t_r]
    \begin{bmatrix}
        (\mathbf{K}^{-1}u)z_i \\
        1 
    \end{bmatrix}
    $
    }
\label{eq:planar-homography-transformation}
\end{equation}
where $u$ denotes the coordinate of a pixel in the support image frame, $\mathbf{K}$ denotes the camera intrinsic matrix, $\{R,t\}$ denotes the rotation and translation of the support image coordinates to the coordinates of the reference image. 

The transformed coordinate mapping is used to construct the warped feature volume, with dimensions of $C \times D \times H' \times W'$, where $C$ denotes the number of channels, and $H'$ and $W'$ denotes the scaled down height and width. When the reference and support feature volumes are concatenated together, we obtain a raw matching volume of $2C \times D \times H' \times W'$ dimension. For $N-1$ support images, we will have $N-1$ raw matching volumes. 3D convolution layers process the raw matching volumes and reduce the dimension to $C \times D \times H' \times W'$. Information across the $N-1$ raw matching volumes are then aggregated through a view average pooling operation. We then apply a series of 3D convolutions to further regularize the aggregated volume to produce the regularized matching volume, which encapsulates the local features for matching.

\textbf{Context volume.} To learn the global 2D context, we apply a ResNet-50 \cite{He2016DeepRL} architecture and obtain a feature volume with dimensions of $D \times H' \times W'$. Notice the number of channels is equivalent to the number of sampled depths $D$ in matching volume. This 2D context volume is then fused with the 3D matching volume through expanding the dimensions to $1 \times D \times H' \times W'$. After concatenation of the regularized matching volume and the 2D context volume, we obtain a final cost volume of size $(C+1) \times D \times H' \times W'$ which encapsulates the local and global context. 

\textbf{Rough depth map.} The cost volume is fed through consecutive convolution layers to obtain volumes with size $D \times H' \times W'$. A softmax operator is applied on the $D$ dimension to create a probability volume $P$, from which the expected depth is extracted using the soft argmax operation.

\subsection{Perception Predictions}

Extraction of high-level perception predictions begins with the concatenation between the low resolution depth map and extracted features of the reference image $I_{reference}$, which enable learning with combined depth, colour, and textual cues, similar to \cite{pmlr-v164-kollar22a,Xie2019TheBO}. The output passes through a ResNet-18-FPN feature backbone before reaching the perception heads.

\textbf{ Pose and 3D bounding box estimation.}
This involves predicting the oriented bounding box of rigid objects, including the translation $t \in \mathbb{R}^{1\times3}$, rotation $R \in \mathbb{R}^{3\times3}$, and size of each object instance along three axis $S \in \mathbb{R}^{3}$. The procedure used for estimating object pose involves several components, including differentiating object instances, deriving object size, and predicting object translation and rotation.

To differentiate object instances, we model each object as a bivariate normal distribution around its center and conduct peak detection. The heatmap can be computed as:
\begin{equation}
    \text{HeatMap}(\mathbf{p}) = \max_{i \in \mathcal{O}}(\mathcal{N}(\mathbf{p} ; \mu_i, \sigma_i))
\end{equation}
where $\mathcal{I}_{reference}$ has pixels $\{\mathbf{p}\}$, $\mathcal{O}$ is the set of object instances within image $\mathcal{I}_{reference}$, and $\mu_i$ and $\sigma_i$ denotes the centroid and covariance of object $i$, respectively.

Determining object size requires prediction of the displacement field, which encapsulates information about the distance between each pixel and the eight vertices of the associated object. The vertex offset can be calculated for an image-plane projected vertex $\mathbf{v}$, and a pixel $\mathbf{p}$ as:
\begin{equation}
    \text{VertexOffset}(\mathbf{p}) = \mathbf{v} - \mathbf{p}
\end{equation}
We operate at the $H/8 \times W/8$ resolution to reduce computation requirements, which results in a displacement field with dimensions of $H/8 \times W/8 \times 16$. To combine displacement fields for all objects within the scene, we consider probabilistic values for each pixel from the heatmap, and merge based on the object with the highest probability. 

To recover the translation, the objects' centroid distances from the camera are regressed as a $H/8 \times W/8$ tensor, which can be used in combination with the camera pose to derive the object translations in world space. Note that the centroid distance field contains information for all objects in the image, the partitioning of which is based on the heatmaps and the object with the highest probability at each pixel, similar to the combined displacement field. Rotation estimation is based on covariance matrix prediction. Computation of the ground-truth covariance of the object begins with sampling points on the object mesh surface in simulation, in the object's local space. The points are then converted to camera space and used to compute the covariance: 
\begin{equation}
    \text{Covariance}(\mathbf{c}_C) = \text{Covariance}(R_W^C \cdot R^W_L \cdot \mathbf{c}_L)
\end{equation}
where $C$ denotes camera frame, $W$ denotes world frame, and $L$ denotes local frame. $\mathbf{c}$ refers to the coordinates of points on the object surface. We then compute covariance on $\mathbf{c}_C$. The ground truth covariance matrix of each object is combined in a similar manner as the displacement field and the centroid distance field, and is used as a supervision signal to enable covariance prediction, which is regressed as a tensor of $H/8 \times W/8 \times 6$, consisting of the elements of the upper triangular matrix of the standard covariance matrix for 3D point clouds. Rotation is then recovered through the Singular Value Decomposition (SVD) of the covariance matrix.

\textbf{Segmentation.} Instance segmentation differentiates the table surface, background, and objects on the table. Training is done using the up-scaling branch approach introduced in \cite{Kirillov2019PanopticFP} with multiple up-sampling layers, where each layer consists of $3 \times 3$ convolution, group norm, ReLU and $2 \times$ bilinear up-sampling.  We use cross entropy loss for training.

\textbf{Depth.} To predict the full resolution depth map, we apply the up-scaling branch similar to  our segmentation head, which enables aggregation of several features across different scales. Training is achieved using the Huber loss to minimize:
\begin{equation}
    L_{depth}= \text{Huber}(f_{depth}(I_{1:N}), D_{reference})
\end{equation}
where $D_{reference}$ denotes the ground truth depth map of the reference frame.

\section{Synthetic Transparent Object Dataset}

Despite recent improvements of real-world data collection, synthetic datasets still lead in terms of throughput, annotation accuracy, object diversity, and scene complexity \cite{duan, dasgupta2021avoe, clemens, tremblay2018falling, duan2021space, johnson2017clevr, baradel2019cophy}. Given the promising performance of stereo vision models \cite{keypose} and sim-to-real training \cite{ClearGrasp, zhu2021rgbd} for transparent object detection, a photo-realistic and large-scale transparent object dataset is needed. We present \dataName, which has wide compatibility with RGB, RGB-D, stereo, and multi-view based methods. For a given scene, we render stereo image pairs from a grid of viewing angles. Additionally, procedural generation of objects and domain randomization of scenes enhance the dataset's complexity and aids model generalization. 

\begin{figure*}[t]
	\centering
	\includegraphics[width=0.75\textwidth]{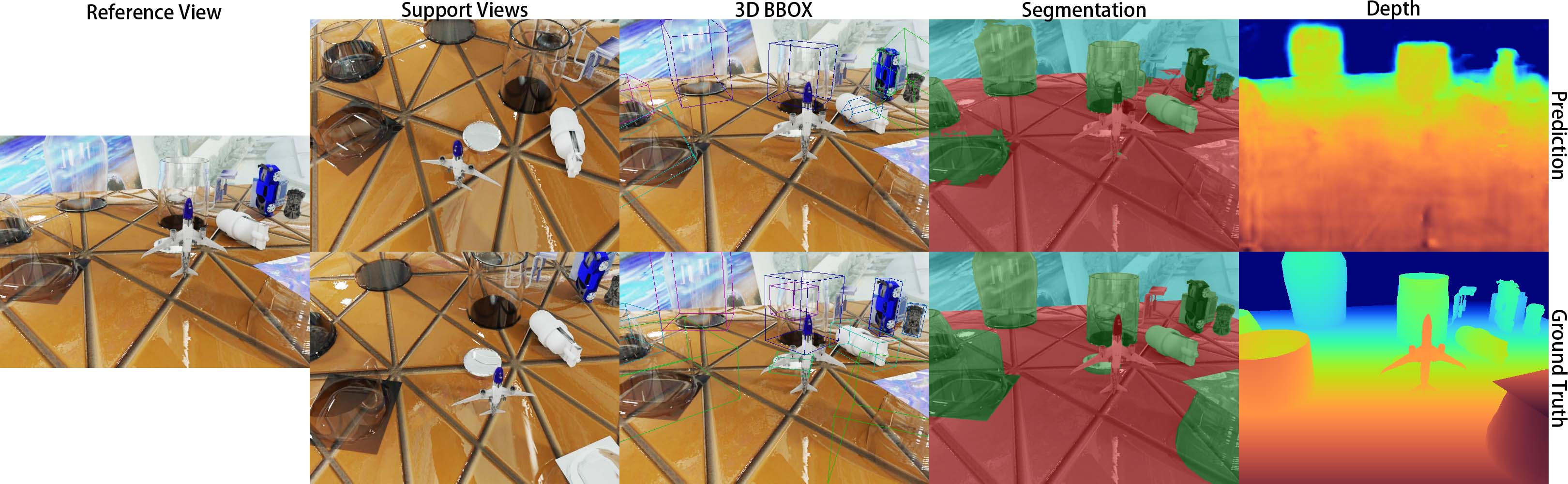} 
	\captionof{figure}{\textbf{Samples and prediction results from \dataName.} (a) Multi-view RGB image (reference view and support views) (b) 3D OBB (GT and prediction)  (c) Instance Segmentation (GT and prediction) (d) Depth (GT and prediction)}
	\label{fig:result}
\end{figure*}

\begin{table}[t]
	\vspace{0pt}
	\centering
	\captionof{table}{\textbf{Transparent Dataset Comparison.} Comparison of \dataName (ours) with ClearGrasp \cite{ClearGrasp}, Trans10K \cite{trans10k}, Keypose \cite{keypose}, TODD \cite{seeingGlass}, ClearPose \cite{clearpose}. Our 3D transparent object dataset has significant advantage over others in terms of sample size, scene complexity, object diversity and annotation richness.
	}
	\resizebox{\linewidth}{!}{
		\begin{tabular}{c|c|c|c|c|c|c}
			\toprule
			\rowcolor[HTML]{CBCEFB}
			$\ $         & \centering Trans10K  & \centering ClearGrasp & \centering TODD &\centering ClearPose & \centering  KeyPose & \centering \textbf{\dataName(Ours)} \cr    
			\midrule
			Samples   & \centering 10K                   & \centering 50K            & \centering 15K              & \centering 360K              & \centering 15K               & \centering 113K  \cr   
			\rowcolor[HTML]{EFEFEF} 
			Objects      & \centering 10K                                     & \centering 10                      & \centering 8                     & \centering 63                      & \centering 10                         & \centering 16K   \cr 
			        
			Scenes       & \centering 10K                                     & \centering 33                      & \centering 22                     & \centering 63                      & \centering 10                         & \centering 1996  \cr         
			        
			\rowcolor[HTML]{EFEFEF} 
			Objs/scene   & \centering 1-20                                   & \centering 1-5                     & \centering 1-3                     & \centering 1-25                     & \centering 1                      & \centering 3-15 \cr          
			        
			RGB          & \centering mono                              & \centering  mono               & \centering mono                 & \centering mono              & \centering  stereo                 & \centering multi-view \cr   
			       
			\rowcolor[HTML]{EFEFEF} 
			Segment & \centering semantic                               & \centering semantic                 & \centering instance               & \centering instance                & \centering instance                   & \centering instance  \cr    
			        
			Depth        & \centering \xmark                                 & \centering \cmark                   & \centering \cmark                 & \centering \cmark                  & \centering \cmark                     & \centering \cmark  \cr      
			        
			\rowcolor[HTML]{EFEFEF} 
			Pose         & \centering \xmark                                 & \centering \xmark                   & \centering \cmark                 & \centering \cmark                  & \centering \cmark                     & \centering \cmark \cr       
			        
			3D Bbox      & \centering \xmark                                 & \centering \xmark                   & \centering \xmark                 & \centering \xmark                  & \centering \xmark                     & \centering \cmark  \cr      
			        
			\rowcolor[HTML]{EFEFEF} 
			Normal       & \centering \xmark                                 & \centering \cmark                   & \centering \xmark                 & \centering \xmark                  & \centering \xmark                     & \centering \cmark  \cr      
			        
			Keypoints    & \centering \xmark                                 & \centering \xmark                   & \centering \xmark                 & \centering \xmark                  & \centering \cmark                     & \centering \cmark  \cr      
			\bottomrule
		\end{tabular}
	}
	
	\label{table:dataset-comparison}
\end{table}

\textbf{Scene Setup.} We use Blender~\cite{blender} for high-fidelity, photo-realistic synthetic data generation \cite{transproteus}. Each scene consists of three parts: background, tabletop, and objects. To diversify scene appearances, we apply domain randomization to select the background from 1000+ High Dynamic Range Image (HDRI) for environments and illumination variances, and the tabletop surface from 1400+ Physics Based Rendering (PBR) materials with varying textures and visual appearances. Multiple light sources are also  introduced at random locations. 

\textbf{Procedural Generation.} Transparent objects are procedurally generated. We employ a method that creates the vessel curvatures using 2D function combinations of linear, polynomial, and sinusoidal functions, with coefficients and parameters differing across vessel. For each generated vessel, we apply a transparent material with randomized properties, including color, index of refraction, transparency, reflection, and roughness, among others. Each scene contains up to seven random transparent objects, with possible occlusion.

\textbf{Object Selection.} As shown in \autoref{fig:dataset_compare}, up to seven generated transparent objects are placed in each scene. Additionally, we randomly place up to eight different objects from a subset of ShapeNet \cite{shapenet} with 13000+ models to simulate occlusion and diversity. The objects are further diversified by randomized scale and orientation. To further mimic a real-life setting, we provide samples in which vessels are filled with varying colors and transparencies of liquid. 

\textbf{Annotations.} The dataset contains a diverse set of annotations and saved scene files. We provide annotations for 57 sets of viewing angles for each scene, where the views are spaced in a grid of an upper-half sphere with random radius. Each view consists of a stereo image pair, and the following annotations are provided for the left image: 2D \& 3D bounding box, object pose, furthest point sampled keypoints, instance segmentation, depth, surface normal, and object centroid heatmap.

\textbf{Dataset Statistics}. As shown in \autoref{table:dataset-comparison}, \dataName consists of 113,772 stereo image pairs of 1996 different scenes containing a combination of 9012 unique opaque objects from ShapeNet \cite{shapenet} and 7010 unique procedural generated transparent objects. \dataName is split into training \& validation sets, with 1575 and 421 scenes each.

\section{Results}
\label{sec:result}
In this section, we analyze the performance of the proposed multi-task model on tasks of depth prediction, segmentation, 3D object detection, and pose estimation. We evaluate and show the robustness of the method against strong baselines using our proposed dataset and a real world dataset \cite{keypose}.

\begin{table}[t]
		\centering
		\caption{\textbf{Pose estimation results on the KeyPose \cite{keypose}.} We assess the performance of DenseFusion, Keypose, SimNet and \modelName with the KeyPose real world dataset for RGB and RGB-D based 6-DoF pose estimation. All models are trained on the KeyPose dataset. DenseFusion is trained twice using raw sensor depth and ground truth depth respectively. Both 2 view and 5 view \modelName  significantly outperform all baseline methods.}
		\resizebox{0.5\textwidth}{!}{
			\begin{tabular}{l|c|c|c|c}
				\toprule
				\rowcolor[HTML]{CBCEFB}
				$\ $                           & modality           & AUC (\textuparrow) & $<2cm$ (\textuparrow) & MAE (\textdownarrow) \\
				
				\midrule
				DenseFusion \cite{densefusion} & RGB-D (truth depth) & 71.9               & 37.5                  & 35.1                 \\
				\rowcolor[HTML]{EFEFEF} 
				DenseFusion \cite{densefusion} & RGB-D (raw depth)   & 63.8               & 18.9                  & 37.2                 \\
				SimNet\cite{pmlr-v164-kollar22a}& stereo RGB        & 87.9              & 83.1                 & 12.6                  \\
				\rowcolor[HTML]{EFEFEF} 
				KeyPose \cite{keypose}         & stereo RGB         & 90.0               & 90.1                  & 9.9                  \\
				\textbf{\modelName(Ours)}              & 2-view RGB     & \textbf{92.7}          & \textbf{93.6}                     & \textbf{7.4}                  \\
				\rowcolor[HTML]{EFEFEF}
				\textbf{\modelName(Ours)}            & 5-view RGB     & \textbf{92.9}             & \textbf{94.0}                     & \textbf{7.2}                    \\
				\bottomrule
			\end{tabular}
			\label{tab:keypose_Metrics}
		}
\end{table}

\subsection{Implementation Details}
\modelName and SimNet are trained on four Nvidia A100 GPUs, with a batch size of 8 to 24 based on view count for 70 epochs on both KeyPose and \dataName datasets. We use the Adam optimizer with $\alpha = 0.0006$, $\beta_1 = 0.9$, and $\beta_2 = 0.99$, and weight decay of 1e-4.

\subsection{Metrics}
We evaluate \modelName for all three of its prediction heads: depth estimation, pose and 3D bounding box estimation and instance segmentation.

For \textbf{depth prediction}, the standard metrics as described in \cite{ClearGrasp} are followed. The prediction and ground truth arrays are first resized to $144 \times 256$ resolution prior to evaluation. Errors are computed using the following metrics, root mean squared error (RMSE), absolute relative difference (REL), and mean absolute error (MAE). 

\begin{table*}[t]
	\centering
	\caption{\textbf{KeyPose Multi-task results.} We train SimNet and \modelName (2/3/5 views) on KeyPose and evaluate their performances in 3D bounding box, 6 DoF pose and segmentation predictions. \modelName has better performance compared to SimNet on KeyPose for all tasks, for both settings.}

	\resizebox{0.7\textwidth}{!}{%
		\begin{tabular}{c|c|c|c|c|c|c|c}
			\toprule
			\rowcolor[HTML]{CBCEFB}
			$\ $  & \multicolumn{2}{c|}{\textbf{3D Bbox}} & \multicolumn{3}{c|}{\textbf{Pose}} & \multicolumn{2}{c}{\textbf{Segmentation}}\\
			\rowcolor[HTML]{CBCEFB}
			$\ $      & 3D mAP (\textuparrow) & 3D IoU (\textuparrow)  & AUC (\textuparrow) & $<2cm$ (\textuparrow) & MAE (mm) (\textdownarrow)  & mAP (\textuparrow) & IoU (\textuparrow)\\
				
			\midrule
			\rowcolor[HTML]{FBE0CB}
			\multicolumn{8}{l}{\textbf{Real Dataset}: Training and Evaluation on KeyPose Dataset}\\
			\midrule
			SimNet\cite{pmlr-v164-kollar22a}&89.40                       &49.80                     &87.89                    &83.14              &12.55                       &99.10                    &92.40       \\
			\rowcolor[HTML]{EFEFEF} 
		\modelName (2 images)   &91.20                       &\textbf{61.40}                       & 92.68                   &93.59           &7.37                       &99.30                    & 92.20       \\
			\modelName (3 images)  &90.40                       & 58.30                      & 92.14                   & 92.76         &    8.00                   &  99.00                  &  90.90         \\
			\rowcolor[HTML]{EFEFEF} 	
			\modelName (5 images)   &\textbf{92.20}                       &60.90                       &\textbf{92.89}                   & \textbf{93.98}          & \textbf{7.15}                      &\textbf{99.80}                    &\textbf{92.80}        \\
			\bottomrule
		\end{tabular}
	}
	\label{tab:keypose_petra_Metrics}
\end{table*}

\begin{table*}[]
	\centering
	\caption{\textbf{\text{\dataName} Multi-task results.} We train SimNet and \modelName (2/3/5 views) on our \dataName dataset, and evaluate depth, 3D bounding box and segmentation prediction performances. \modelName has better performance compared to SimNet on \dataName dataset for all tasks.}

	\resizebox{0.7\textwidth}{!}{%
		\begin{tabular}{c|c|c|c|c|c|c|c}
			\toprule
			\rowcolor[HTML]{CBCEFB}
			$\ $ & \multicolumn{3}{c|}{\textbf{Depth}} & \multicolumn{2}{c|}{\textbf{3D Bbox}} & \multicolumn{2}{c}{\textbf{Segmentation}}\\
			\rowcolor[HTML]{CBCEFB}
			$\ $                             & RMSE (\textdownarrow) & MAE (\textdownarrow) & REL (\textdownarrow) & 3D mAP (\textuparrow) & 3D IoU (\textuparrow)  & mAP (\textuparrow) & IoU (\textuparrow)\\
				
			\midrule
			\rowcolor[HTML]{FBE0CB}
			\multicolumn{8}{l}{\textbf{Synthetic Dataset}: Training and Evaluation on \dataName  Dataset} \\
			\midrule
			SimNet\cite{pmlr-v164-kollar22a} & 1.229            &         1.020             &               0.975                 & 4.65                       &  34.92                      &   48.21                  &      50.52 \\
			\rowcolor[HTML]{EFEFEF} 
			\modelName (2 images)     & 0.134                    & 0.089                    &0.135                                & 40.79                     &  45.95                      & 84.94                   & 79.52        \\
			\modelName (3 images)    &   0.125                    & 0.083                     & 0.125                               &  42.53                    &46.17                       &     \textbf{ 87.75 }               &\textbf{81.89}       \\
			\rowcolor[HTML]{EFEFEF} 	
			\modelName (5 images)   &   \textbf{0.124}                    &   \textbf{0.080}                    & \textbf{0.117}                                &   \textbf{46.99}                    &   \textbf{48.44}                     &    87.24                  & 81.30     \\
			\bottomrule
		\end{tabular}
	}
	\label{tab:petra_Metrics}
\end{table*}

For \textbf{6-DoF pose}, Area Under the Curve (AUC),  percentage of
3D keypoint errors $<2cm$ and Mean Absolute Error (MAE). AUC percentage is calculated
based on an X-axis range from 0 to 10 cm, where the curve shows the cumulative percentage of errors under that metric value.

For \textbf{3D bounding box}, 3D intersection over union (3D IoU) is used to measure box fit, and 3D mean average precision (3D mAP) is calculated using 3D IoU $>$ 0.25 as criteria, which is correlated with grasp success rate as shown in \cite{pmlr-v164-kollar22a}.

For \textbf{instance segmentation}, intersection over union (IoU) and mean average precision (mAP) are used to evaluate the predicted mask. For mAP, IoU $>$ 0.5 is used as the threshold.

\subsection{Experiment 1: Multi-view and RGB-D Comparisons}
For transparent objects, the raw depth captured by commodity RGB-D sensors is incomplete and distorted, which naturally form the depth completion task when using RGB-D based methods. However, stereo and multi-view based models do not rely on depth information, thus is advantageous for transparent object related 3D tasks.
To demonstrate the claimed advantage and our \modelName's pose prediction capability, as shown in \autoref{tab:keypose_Metrics}, we compare the pose predicted by RGB-D based DenseFusion \cite{densefusion} with \modelName trained on KeyPose \cite{keypose} dataset. \modelName's two and five view versions both outperform DenseFusion, regardless of whether it is trained on raw distorted depth, or ground truth depth.

\subsection{Experiment 2: Multi-view and Stereo Vision Comparisons}
We conduct experiments to test \modelName against and other stereo RGB based methods. First, we focus on the pose estimation task, where \modelName is trained on the KeyPose \cite{keypose} dataset and compared with current state-of-the-art networks. The results are listed in \autoref{tab:keypose_Metrics}. For the baselines, KeyPose \cite{keypose} and SimNet \cite{pmlr-v164-kollar22a} are both stereo image based models, where KeyPose predicts object pose by keypoints, and SimNet is a multitask model with a pose estimation branch. \modelName takes two views of the scene as input to match the amount of information received by RGB-D and stereo methods. \modelName demonstrates a significant advantage in all three pose estimation metrics when compared to all baselines. 

\subsection{Experiment 3: Evaluating Multi-task Performance}
 We conduct experiments to study \modelName's multi-task capabilities when compared against the previous SOTA method \cite{pmlr-v164-kollar22a} across depth estimation, 3D orientated bounding box or pose estimation, and segmentation tasks. Our evaluations are done on two datasets, namely KeyPose \cite{keypose}, and \dataName.  For each experiment, all models are trained and evaluated on the same corresponding dataset.  For simple scenes, including single object scenes from KeyPose, as shown in \autoref{tab:keypose_petra_Metrics}, \modelName outperforms SimNet \cite{pmlr-v164-kollar22a} with all experimented view counts, including two views. This shows the advantage of multi-view over stereo vision without increasing view count. For complex settings, including multiple objects from \dataName, as shown in \autoref{tab:petra_Metrics}, \modelName has better performance compared to SimNet by a significant margin. However, 3D bounding box and pose estimation for complex scenes and novel objects presented in \dataName remains challenging given lower results for all methods, in comparison to experimental results on the KeyPose dataset.

\subsection{Ablation Study} 
To quantitatively evaluate the performance gain from increasing image views, \modelName with view counts of two, three, and five are trained. We present the evaluation of their performances on both Keypose and \dataName datasets, in \autoref{tab:keypose_petra_Metrics} and \autoref{tab:petra_Metrics}, respectively. For both datasets, \modelName has superior performance over the baseline stereo SimNet method, and increasing view count generally yields improved results, especially for harder tasks, for example, the 3D bounding box estimation result for the complex \text{\dataName} dataset. For simpler tasks like segmentation, we see that increasing the view count leads to marginal improvement, the reason for this is because segmentation is a 2D task, and hence will not benefit as much from the richer 3D information that additional views provide. Overall, we see that there is still room for improvement for all metrics, which reveals the challenging nature of the dataset.

\section{Conclusion}

In this work, we proposed a large-scale photo-realistic multiview dataset, \dataName, for pre-training multiview networks, in addition to a novel end-to-end multiview-based method for multi-task learning, \modelName. We evaluate the performance of \modelName on both synthetic and real datasets, including \dataName (Ours) and  KeyPose Dataset \cite{keypose}, and observe that we outperform previous baselines by a large margin in depth estimation, segmentation, and scene understanding (3D bounding box and pose estimation). Future directions worth exploring include sim-to-real transfer using the large-scale photo-realistic dataset, and leveraging the perception predictions for scene-graph and/or grasp generation for downstream manipulation and planning tasks. We hope this work can help accelerate future research in household manipulation and laboratory automation.

\section*{ACKNOWLEDGMENT}
AG \& AAG are CIFAR AI Chairs. AAG is a Lebovic Fellow. AG and FS are also supported in part through the NSERC Discovery Grants Program. The authors would like to acknowledge Vector Institute and Compute Canada for computing services. AAG and HX thank the Canada 150 Research Chair funding from NSERC, Canada. AAG is thankful for the generous support of Dr. Anders G. Fr\o seth. The authors would like to thank Kourosh Darvish for constructive feedback and discussions on the manuscript.


\newpage
\renewcommand*{\bibfont}{\small}
\bibliographystyle{IEEEtran}
\bibliography{0-petra}

\newpage

\end{document}